\documentclass[10pt,twocolumn,letterpaper]{article}

\usepackage{cvpr}
\usepackage{times}
\usepackage{epsfig}
\usepackage{graphicx}
\usepackage{amsmath}
\usepackage{amssymb}

\usepackage{times}
\usepackage{epsfig}
\usepackage{graphicx}
\usepackage{amsmath}
\usepackage{amssymb}
\usepackage{subcaption}
\usepackage{multirow}
\usepackage{booktabs}
\usepackage{mathtools}
\usepackage{algorithm}
\usepackage{algpseudocode}
\usepackage{anyfontsize}
\usepackage{t1enc}
\usepackage[table]{xcolor}
\usepackage{enumitem}
\usepackage{bbding}
\usepackage{subcaption}
\usepackage[pagebackref=true,breaklinks=true,letterpaper=true,colorlinks,bookmarks=false]{hyperref}

\cvprfinalcopy 

\def\cvprPaperID{2032} 

\ifcvprfinal\fi
\begin{document}

\title{Tell-and-Answer: Towards Explainable Visual Question \\ Answering using Attributes and Captions}

\author{Qing Li\textsuperscript{1}, Jianlong Fu\textsuperscript{2}, Dongfei Yu\textsuperscript{1}, Tao Mei\textsuperscript{2}, Jiebo Luo\textsuperscript{3}\\
\textsuperscript{1}University of Science and Technology of China\\
\textsuperscript{2}Microsoft Research, Beijing, China\\
\textsuperscript{3}University of Rochester, Rochester, NY\\
{\tt\small \textsuperscript{1}\{sealq, ydf2010\}@mail.ustc.edu.cn, \textsuperscript{2}\{jianf, tmei\}@microsoft.com, \textsuperscript{3}jluo@cs.rochester.edu}
}

\maketitle

\begin{abstract}
	Visual Question Answering (VQA) has attracted attention from both computer vision and natural language processing communities. Most existing approaches adopt the pipeline of representing an image via pre-trained CNNs, and then using the uninterpretable CNN features in conjunction with the question to predict the answer. Although such end-to-end models might report promising performance, they rarely provide any insight, apart from the answer, into the VQA process. In this work, we propose to break up the end-to-end VQA into two steps: \textbf{explaining} and \textbf{reasoning}, in an attempt towards a more explainable VQA by shedding light on the intermediate results between these two steps. To that end, we first extract attributes and generate descriptions as explanations for an image using pre-trained attribute detectors and image captioning models, respectively. Next, a reasoning module utilizes these explanations in place of the image to infer an answer to the question. The advantages of such a breakdown include: (1) the attributes and captions can reflect what the system extracts from the image, thus can provide some explanations for the predicted answer; (2) these intermediate results can help us identify the inabilities of both the image understanding part and the answer inference part when the predicted answer is wrong. We conduct extensive experiments on a popular VQA dataset and dissect all results according to several measurements of the explanation quality. Our system achieves comparable performance with the state-of-the-art, yet with added benefits of explanability and the inherent ability to further improve with higher quality explanations.
\end{abstract}

\vspace{-0.5cm}
\section{Introduction}
Answering textual questions from images, which is referred to as visual question answering, presents fundamental challenges to both computer vision and natural language processing communities. This task is considered as the milestone of ``AI-complete.'' Compared with text-based applications (e.g., sentence generation \cite{sutskever2011generating} and text QA \cite{weston2015towards}), VQA takes one step further, requiring a machine to be equipped with cross-modality understanding across language and vision \cite{antol2015vqa,zhu2016visual7w,goyal2017making}.

\begin{figure}[!tb]
	\centering {\includegraphics[width=0.5\textwidth]{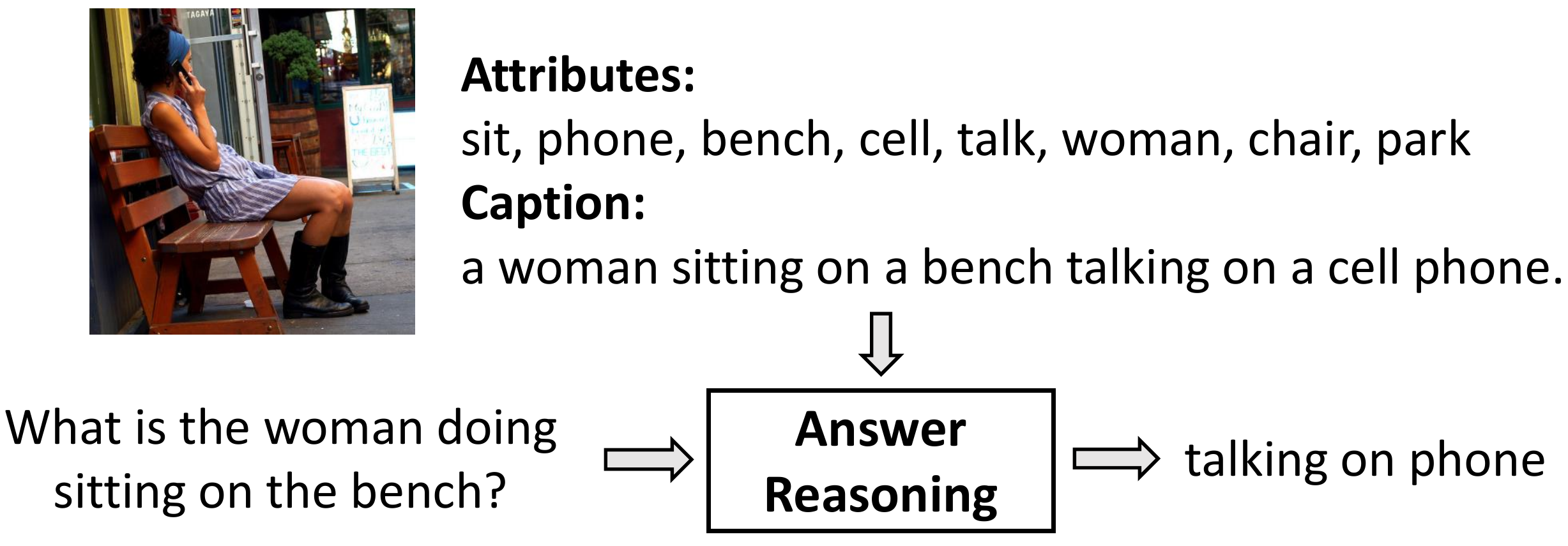}}
	\caption{An example of explanation and reasoning in VQA. We first extract attributes in the image such as ``sit'', ``phone'' and ``woman.'' A caption is also generated to encode the relationship between these attributes, e.g. ``woman sitting on a bench.'' Then a reasoning module uses these explanations to predict an answer ``talking on the phone.''}
	\label{fig:intro_case}
\end{figure}

\begin{figure}[!tb] \vspace{-2mm}
	\centering {\includegraphics[width=0.5\textwidth]{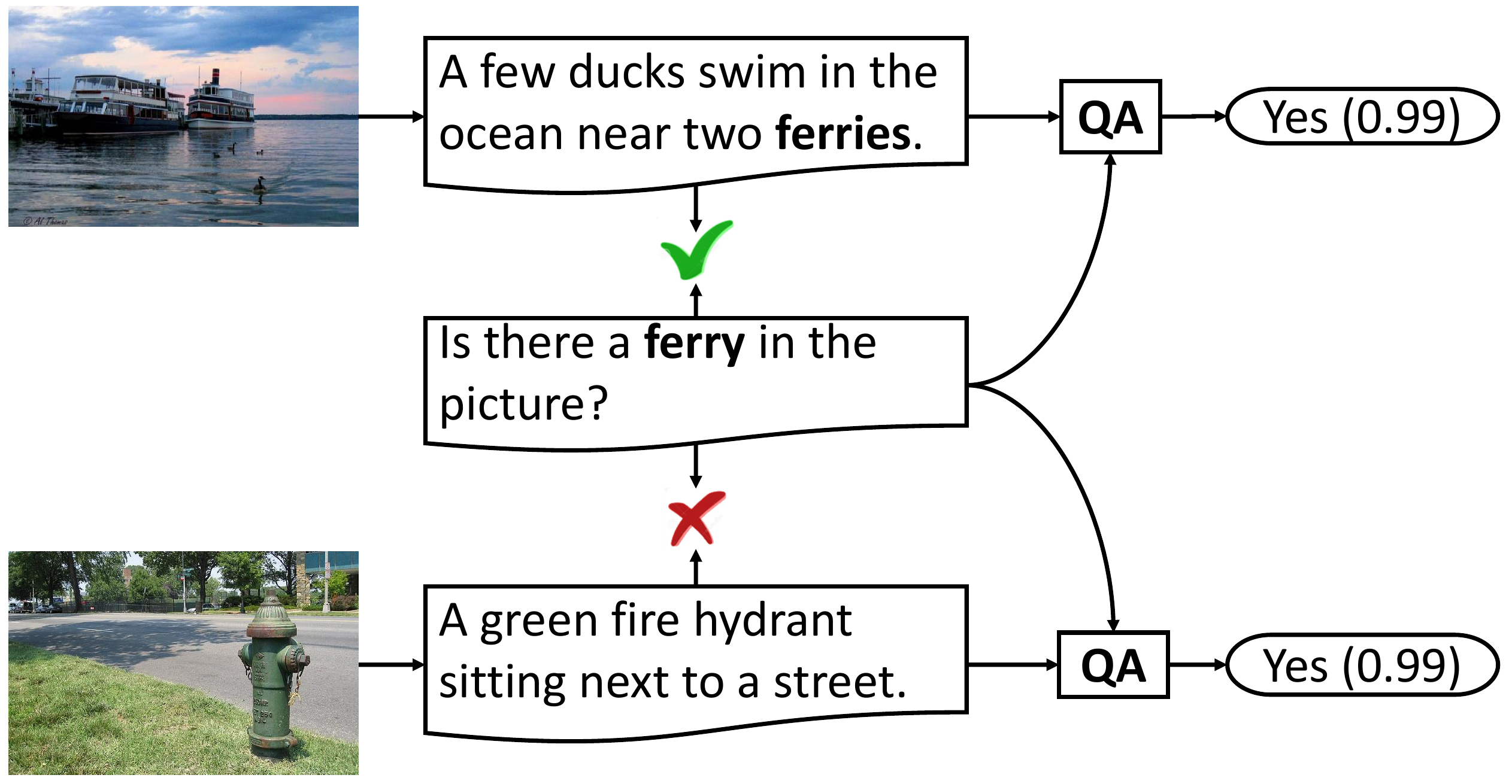}}
	\caption{Two contrasting cases that show how the explanations can be used to determine if the system learns from a training set bias or from the image content.}
	\label{fig:intro_case_2}
\end{figure}

Significant progress has been made on VQA in recent years \cite{wu2016value,yang2016stacked,noh2016image,xiong2016dynamic}. A widely used pipeline is to first encode an image with Convolutional Neural Networks (CNNs) and represent associated questions with Recurrent Neural Networks (RNNs), and then formulate the vision-to-language task as a classification problem on a list of answer candidates. Although promising performance has been reported, this end-to-end paradigm fails to provide any insight to illuminate the VQA process. In most cases, giving answers without any explanation cannot satisfy human users, especially when the predicted answer is not correct.
More frustratingly, the system gives no hint about which part of such systems is the culprit for a wrong answer.

To address the above issues, we propose to break up the popular end-to-end pipeline into two steps: \textbf{explaining} and \textbf{reasoning}. The philosophy behind such a break-up is to mimic the image question answering process of human beings: first understanding the content of the image and then performing inference about the answer according to the  understanding.
As is shown in Fig.\ref{fig:intro_case}, we first generate two-level explanations for an image via pre-trained attribute detectors and image captioning model: 1). word-level: attributes, indicating individual objects and attributes the system learns from the image. 2). sentence-level: captions, representing the relationship between the objects and attributes. Then the generated explanations and question are infused to a reasoning module to predict an answer. The reasoning module is mainly composed of LSTMs. 

Our method has three benefits.
First, these explanations are interpretable. According to the attributes and captions, we can tell what objects, attributes and their relationship the machine learns from the image as well as what information is lost during the image understanding step. In contrast, the fully-connected layer features of CNNs are usually uninterpretable to humans. When the predicted answer is correct, these attributes and captions can be provided for users as the supplementary explanations to the answer.
Second, the separation of explaining and reasoning enables us to localize which step of the VQA process the error comes from when the predicted answer is wrong. If the explanations don't include key information to answer the question, the error is caused by missing information during the explaining step. Otherwise, the reasoning module should be responsible for the wrong answer.
Third, the explanations can also indicate whether the system really finds key information from the image to answer the question or merely \textit{guesses} an answer. This especially works against a learner simply taking advantage of the training set bias instead of learning from the actual image content. Fig.\ref{fig:intro_case_2} presents two contrasting cases to illustrate this. In the first case, both the generated caption and the question include the key concept ``ferry'', so the answer ``Yes'' with a high probability is reliable. However, although the answer ``Yes'' has the same high probability in the second case, the caption is irrelevant to the question. This implies that the system sticks to a wrong answer even with the correct input from sentence generation. A further investigation suggests that the error is due to the training set bias. A large proportion of questions starting with ``is there'' in the training set have the answer ``Yes''. 

To our knowledge, this is the first effort to break down the previous end-to-end pipeline to shed light on the VQA process. Our main contributions are summarized as follows:
\begin{itemize}[nosep]
	\item We propose to formulate VQA into two separate steps: \textbf{explaining} and \textbf{reasoning}. Our framework generate attributes and captions for images to shed light on why the system predicts any specific answer.
	\item We adopt several ways to measure the quality of explanations and demonstrate a strong correlation between the explanation quality and the VQA accuracy. The current system achieves comparable performance to the state-of-the-art and can naturally improve with explanation quality.
	\item Extensive experiments are conducted on the popular VQA dataset \cite{antol2015vqa}. We dissect all results according to the measurements of the quality of explanations to present a thorough analysis of the strength and weakness of our framework.
\end{itemize}

\section{Related Work} \label{sec:RW}
There is a growing research interest in the task of visual question answering. In this section, we first discuss the popular CNN-RNN paradigm used in VQA and then summarize the recent advances from two directions.

\noindent\textbf{CNN-RNN}. Inspired by the significant progress of image captioning achieved by combining CNN and RNN, the paradigm of CNN-RNN has become the most common practice in VQA \cite{gao2015you,malinowski2015ask,ren2015exploring}. Visual features of images are extracted via pre-trained convolutional neural networks (CNNs). Different from image captioning models, recurrent neural networks (RNNs) in VQA are used to encode questions.
\cite{ren2015exploring} treat the image as the first token and feed it into RNN along with the question to predict an answer.
Inspired by \cite{donahue2015long}, \cite{malinowski2015ask} pass the image into RNN at each time step, instead of only seeing the image once.
\cite{gao2015you} adapt m-RNN models \cite{mao2014deep} to handle the VQA task in the multi-lingual setting.
Despite these methods show promising results, they fail to recognize novel instances in images and are highly relied on questions while neglecting the image content \cite{agrawal2016analyzing,jabri2016revisiting}.

\noindent\textbf{Attention in VQA.} The attention mechanism is firstly used in the machine translation task \cite{bahdanau2014neural} and then is brought into the vision-to-language tasks \cite{xu2015show,you2016image,xu2016ask,yang2016stacked,lu2016hierarchical,ilievski2016focused,nam2017dual,yu2017multi}. The visual attention in the vision-to-language tasks is used to address the problem of ``where to look'' \cite{shih2016look}. 
In VQA, the question is used as a query to search for the relevant regions in the image.
\cite{yang2016stacked} propose a stacked attention model which queries the image for multiple times to infer the answer progressively.
Beyond the visual attention, Lu \textit{et al.} exploit a hierarchical question-image co-attention strategy to attend to both related regions in the image and crucial words in the question. 
\cite{nam2017dual} propose the dual attention network, which refines the visual and textual attention via multiple reasoning steps.
\cite{fukui2016multimodal} incorporate a powerful feature fusion method into visual attention and obtain impressive results. 
Attention mechanism can find the question-related regions in the image, which can account for the answer to some extent.
But the attended regions still don't explicitly exhibit what the system learns from the image and it is also not explained why these regions should be attended to.

\noindent\textbf{High-level Concepts.} In the scenario of vision-to-language, high-level concepts exhibit superior performance than the low-level or middle-level visual features of the image \cite{fang2015captions,wu2016value,wu2016ask}.
Each concept corresponds to a word mined from the training image captions and represents the objects and attributes presented in the image.
\cite{fang2015captions} first learn independent detectors for visual words based on a multi-instance learning framework and then generate descriptions for images based on the set of visually detected words via a maximum entropy language model.
\cite{wu2016value} presents a thorough study on how much the high-level concepts can benefit the image captioning and visual question answering tasks.
\cite{wu2016ask} leverage the high-level concepts and captions to search external knowledge bases to further improve the VQA performance.
These work mainly uses high-level concepts to obtain a better performance.
Different from these work, our paper is focused on fully exploiting the readability and understandability of attributes and captions to explain the process of visual question answering and use these explanations to analyze our system.

\begin{figure*}[!tb] \vspace{-10mm}
	\centering {\includegraphics[width=1\textwidth]{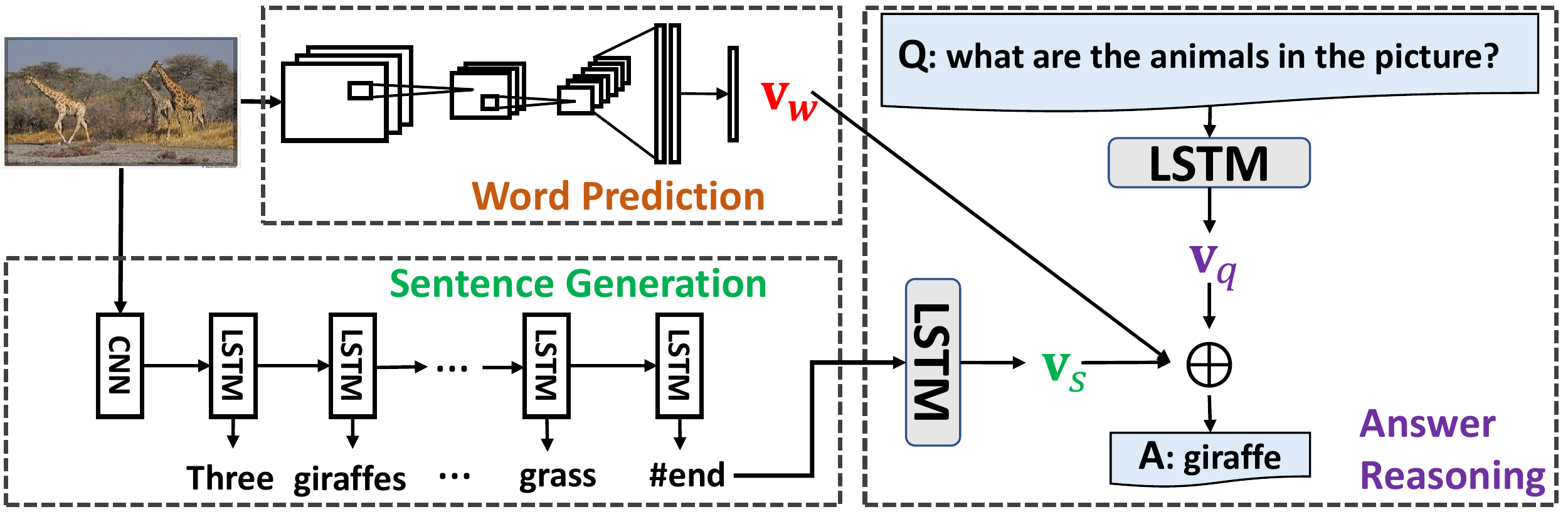}}
	\caption{An overview of the proposed framework for VQA with three modules: word prediction (upper left), sentence generation (lower left), answer reasoning (right). \textbf{Explaining}: in word prediction, the image is fed into pre-trained visual detectors to extract word-level explanation, which is represented by probability vector $\mathbf{v}_w$; in sentence generation, we input the image to pre-trained captioning model to generate a sentence-level explanation. \textbf{Reasoning}: the caption and question are encoded by two different LSTMs into $\mathbf{v}_s$ and $\mathbf{v}_q$, respectively. Then $\mathbf{v}_q, \mathbf{v}_w$ and $\mathbf{v}_s$ are concatenated and fed to a fully connected layer with softmax to predict an answer.}
	\label{fig:framework}
\end{figure*}

\section{Methodology} \label{sec:ME}
In this section, we introduce the proposed framework for the breakdown of VQA. As illustrated in Figure \ref{fig:framework}, the framework consists of three modules: word prediction, sentence generation, and answer reasoning. Next, we describe the three modules in details.

\subsection{Word Prediction}
From the work \cite{fang2015captions,wu2016value,you2016image}, we have learned that explicit high-level attributes can benefit vision-to-language tasks. In fact, besides performance gain, the readability and understandability of attributes also makes them an intuitive way to explain what the model learns from images.

Similar to \cite{fang2015captions}, we first build a word list based on MS COCO Captions \cite{chen2015microsoft}. We extract the most $N$ frequent words in all captions and filter them by lemmatization and removing stop words to determine a list of 256 words, which cover over 90\% of the word occurrences in the dataset. These words can be any part of speech, including nouns (object names), verbs (actions) or adjectives (properties). In contrast to \cite{fang2015captions}, our words are not tense or plurality sensitive, for example, ``horse'' and ``horses'' are considered as the same word. This significantly decreases the size of our word list.
Given the word list, every image is paired with multiple labels (words) according to its captions.
Then we formulate word prediction as a multi-label classification task as \cite{wu2016value}.

\begin{figure}[!tb]
	\centering {\includegraphics[width=0.45\textwidth]{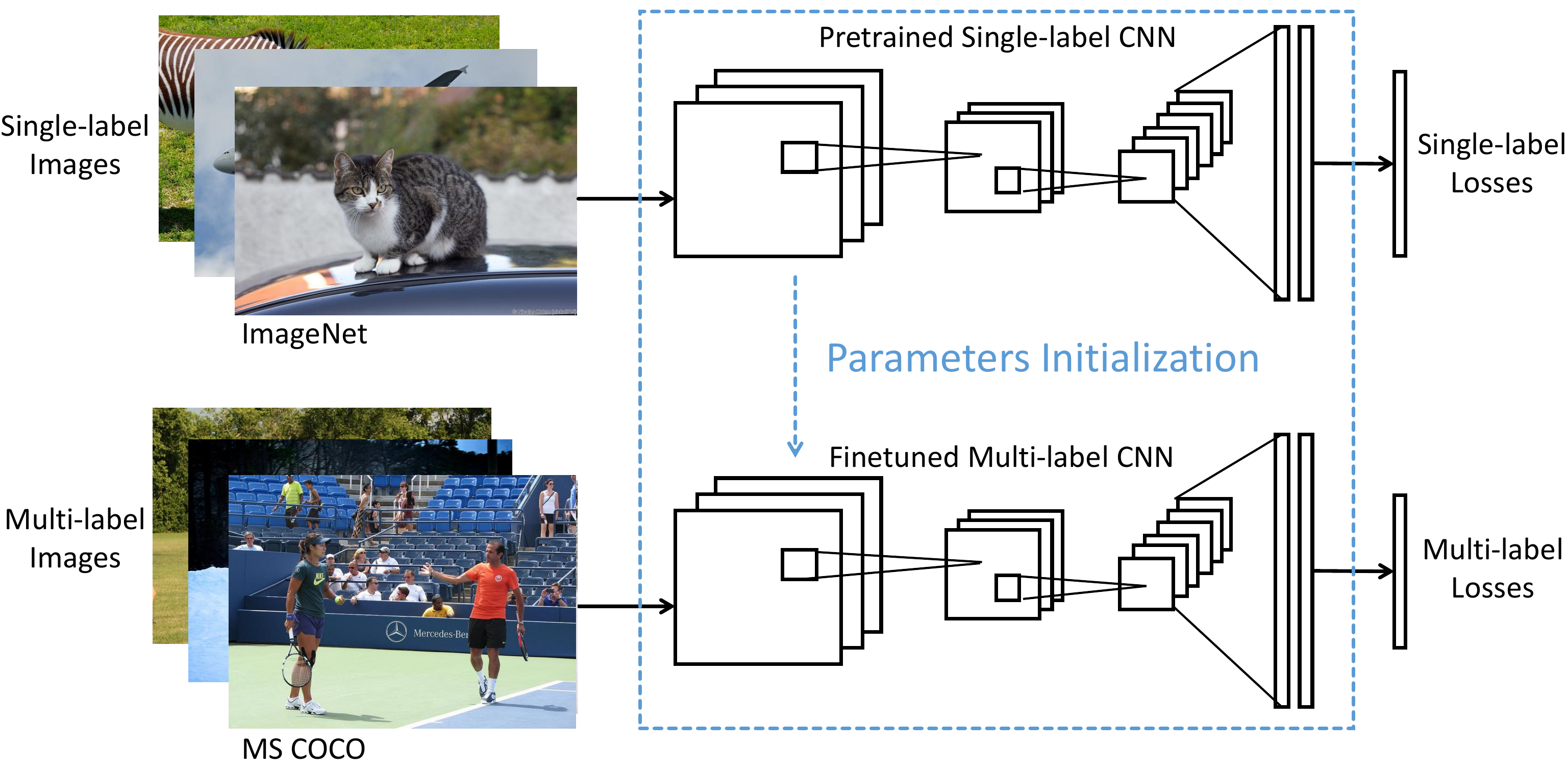}}
	\caption{Word prediction CNN: the model is firstly initialized from ResNet-152 pre-trained on ImageNet. Then the model is fine-tuned on our image-words dataset built from MS COCO captions.}
	\label{fig:word_prediction}
\end{figure}

Figure \ref{fig:word_prediction} summarizes our word prediction network. Different from \cite{wu2016value}, which uses VggNet \cite{simonyan2014very} as the initialization of the CNN, we adopt the more powerful ResNet-152 \cite{he2016deep} pre-trained on ImageNet \cite{deng2009imagenet}. After initialization, the CNN is fine-tuned on our image-words dataset by minimizing the element-wise sigmoid cross entropy loss:
\begin{equation}
J = \frac{1}{N}\sum_{i=1}^{N}\sum_{j=1}^{V}-y_{ij}\log p_{ij}-(1-y_{ij})\log (1-p_{ij})
\end{equation}
where $N$ is batch size, $V$ is the size of word list, $\mathbf{y}_i=[y_{i1}, y_{i2},...,y_{iV}], y_{ij}\in\{0,1\}$ is the label vector of the $i^{th}$ image, $\mathbf{p}_i=[p_{i1}, p_{i2},...,p_{iV}]$ is the probability vector.

In the testing phase, instead of using region proposals like \cite{wu2016value}, we directly feed the whole image into the word prediction CNN in order to keep simple and efficient. As a result, each image is encoded into a fixed-length vector, where each dimension represents the probability of the corresponding word occurring in the image.

\noindent\textbf{Word Quality Evaluation.}
We adopt two metrics to evaluate the predicted words. The first measures the \textbf{accuracy} of the predicted words by computing cosine similarity between the label vector $\mathbf{y}$ and the probability vector $\mathbf{p}$: 
\begin{equation} \label{eq:word accuracy}
a = \frac{\mathbf{y}^T\mathbf{p}}{||\mathbf{y}||\cdot||\mathbf{p}||}
\end{equation}
However, this metric disregards the extent to which the predicted words are relevant to the question. Intuitively speaking, question-relevant explanations for images should be more likely to help predict right answers than irrelevant ones. Therefore, we propose another metric to measure the \textbf{relevance} between the words and the question. We first encode the question into a 0-1 vector $\mathbf{q}$ in terms of the word list. Then the relevance is computed as:
\begin{equation} \label{eq:w-q relevance}
r = \frac{\mathbf{q}^T\mathbf{p}}{||\mathbf{q}||\cdot||\mathbf{p}||}
\end{equation}

\subsection{Sentence Generation}
This section we talk about generating sentence-level explanations for images by using a pre-trained image captioning model.
Similar to \cite{vinyals2015show}, we train an image captioning model by maximizing the probability of the correct caption given an image.
Suppose we have an image $\mathcal{I}$ to be described by a caption $\mathcal{S} = \{s_1, s_2, ...,s_L\}, s_t\in\mathcal{V}$, where $\mathcal{V}$ is the vocabulary, $L$ is the caption length.
First the image $\mathcal{I}$ is represented by the activations of the first fully connected layer of ResNet-152 pre-trained on ImageNet, denoted as $\mathbf{v}_i$.
The caption $\mathcal{S}$ can be represented as a sequence of one-hot vector $\mathbf{S}=\{\mathbf{s}_1, \mathbf{s}_2, ...,\mathbf{s}_L\}$.
Then we formulate the caption generation problem as minimizing the cost function:
\begin{equation}
\begin{split}
J(\mathbf{v}_i,\mathbf{S})     &= -\log P(\mathbf{S}|\mathbf{v}_i) \\
&= -\sum_{t=0}^{L}{\log P(\mathbf{s}_t|\mathbf{v}_i, \mathbf{s}_1, ...,\mathbf{s}_{t-1})}
\end{split}
\end{equation}
where $P(\mathbf{s}_t|\mathbf{v}_i, \mathbf{s}_1, ...,\mathbf{s}_{t-1})$ is the probability of generating the word $\mathbf{s}_t$ given the image representation  $\mathbf{v}_i$ and previous words $\{\mathbf{s}_1, ...,\mathbf{s}_{t-1}\}$. We employ a single-layer LSTM with 512-dimensional hidden states to model this probability. In the testing phase, the image is input to pre-trained image captioning model to generate sentence-level explanation.

\noindent\textbf{Sentence Quality Evaluation.}
Similar to word quality evaluation, we evaluate the quality of the generated sentence from two perspectives: \textbf{accuracy} and \textbf{relevance}. The former one is an average fusion of four widely used metrics: BLEU@N, METEOR, ROUGE-L and CIDEr-D, which try to consider the accuracy of the generated sentence from different perspectives. Detailed explanations about these metrics can be seen in \cite{chen2015microsoft}. Note that we normalize all the metrics into $[0,1]$ before fusion.
The latter metric is to measure the relevance between the generated sentence and the question. The binary TF weights are calculated over all words of the sentence to produce an integrated representation of the entire sentence, denoted by $\mathbf{s}$. Likewise, the question can be encoded to $\mathbf{q}$. The relevance is computed as:
\begin{equation}
r = \frac{\mathbf{q}^T\mathbf{s}}{||\mathbf{q}||\cdot||\mathbf{s}||}
\end{equation}
\subsection{Answer Reasoning}
This section we discuss the reasoning module. Suppose we have an image $\mathcal{I}$ explained by the predicted words $\mathcal{W}$ and the generated sentence $\mathcal{S}$, the question $\mathcal{Q}$ and the answer $\mathcal{A}$. As shown in Fig.\ref{fig:framework}, we denote the representations of the predicted words $\mathcal{W}$ as $\mathbf{v}_s$. The caption $\mathcal{S}$ and question $\mathcal{Q}$ are encoded by two different LSTMs into $\mathbf{v}_s$ and $\mathbf{v}_q$, respectively. What bears mentioning is that these two LSTMs share a common word-embedding matrix, but not other parameters, because the question and caption have different grammar structures and similar vocabularies.
At last, the $\mathbf{v}_w$, $\mathbf{v}_s$, and $\mathbf{v}_q$ are concatenated and fed into a fully connected layer with softmax to predict the probability on a set of candidate answers:
\begin{align}
\mathbf{v} &= [\mathbf{v}_w^T~~\mathbf{v}_s^T~~\mathbf{v}_q^T]^T \label{eq:concate} \\
\mathbf{p} &= softmax(\mathbf{Wv+b}) 
\end{align}
where $\mathbf{W,b}$ are the weight matrix and bias vector of the fully connected layer.
The optimizing objective for the reasoning module is to minimize the cross entropy loss as:
\begin{equation} \label{eq:E_VQA}
J(\mathcal{I,Q,A}) = J(\mathcal{W,S,Q,A}) = -\log \mathbf{p}(\mathcal{A})
\end{equation}
where $\mathbf{p}(\mathcal{A})$ denotes the probability of the ground truth $\mathcal{A}$.

\section{Experiments and Analysis} \label{sec:EA}
\subsection{Experiment Setting}
\noindent\textbf{Dataset.} We evaluate our framework on VQA-real \cite{antol2015vqa} dataset. For each image in VQA-real, 3 questions are annotated by different workers and each question has 10 answers from different annotators. We follow the official split and report our results on the open-ended task.

\noindent\textbf{Metric.} We use the accuracy metric provided by \cite{antol2015vqa}: min($\frac{\text{\#humans giving that answer}}{3}$, 1) , i.e., an answer is deemed 100\% accurate if at least three workers provided that exact answer.

\noindent\textbf{Ablation Models.} To analyze the contribution of word-level and sentence-level explanations, we ablate the full model and evaluate several variants as:
\begin{itemize}
	\item \textbf{Word-based VQA}: use the feature concatenation of the predicted words and question in Eq.\ref{eq:concate}.
	\item \textbf{Sentence-based VQA}: use the feature concatenation of the generated sentence and question in Eq.\ref{eq:concate}.
	\item \textbf{Full VQA}: use the feature concatenation of words, sentence, and question in Eq.\ref{eq:concate}.
\end{itemize}

\noindent\textbf{Model Configuration.}
We explain our model setting and training details here.
We build the vocabulary from questions in VQA-real \cite{antol2015vqa}. We convert all sentences to lower case and tokenize them by NLTK tool and discard words which occur less than five times, resulting in the final vocabulary with 6,148 unique words. For the question and caption encoders in answer reasoning module, we utilize single-layer LSTMs with 512-dimensional hidden states and word embedding size is 512. We select the most frequent 3000 answers in training set as the list of candidate answers.
In training, we use Adam solver \cite{kingma2014adam} with a batch size of 128. The initial learning rate is 0.01 and is dropped to 0.001 after first 10 epochs. The training is stopped after another 10 epochs. In addition, dropout \cite{srivastava2014dropout} and batch normalization \cite{ioffe2015batch} are used in the training procedure.

\subsection{Word-based VQA}
\begin{table}[!htb] \vspace{-5mm}
	\caption{The relationship between word quality and VQA accuracy (\%).}
	\begin{subtable}{.5\linewidth}
		\centering
		\caption{Word accuracy}
		\label{tab:word_accuracy}%
		\begin{tabular}{cc}
			\toprule
			\begin{tabular}{@{}c@{}}Word \\ accuracy\end{tabular}
			& 
			\begin{tabular}{@{}c@{}}VQA \\ accuracy \end{tabular} \\
			\midrule
			\relax [0.0, 0.2) & 46.30 \\
			\relax [0.2, 0.8) & 55.84 \\
			\relax [0.8, 1.0) & 58.52 \\
			\bottomrule
		\end{tabular}
	\end{subtable}%
	\begin{subtable}{.5\linewidth}
		\centering
		\caption{W-Q Relevance}
		\label{tab:w-q relevance}%
		\begin{tabular}{cc}
			\toprule
			\begin{tabular}{@{}c@{}}W-Q \\ Relevance\end{tabular}
			& 
			\begin{tabular}{@{}c@{}}VQA \\ accuracy \end{tabular} \\
			\midrule
			\relax [0.0, 0.2) & 54.69 \\
			\relax [0.2, 0.8) & 60.23 \\
			\relax [0.8, 1.0) & 76.15 \\
			\bottomrule
		\end{tabular}
	\end{subtable} 
\end{table}

\begin{figure}[!tb] \vspace{-3mm}
	\centering {\includegraphics[width=0.45\textwidth]{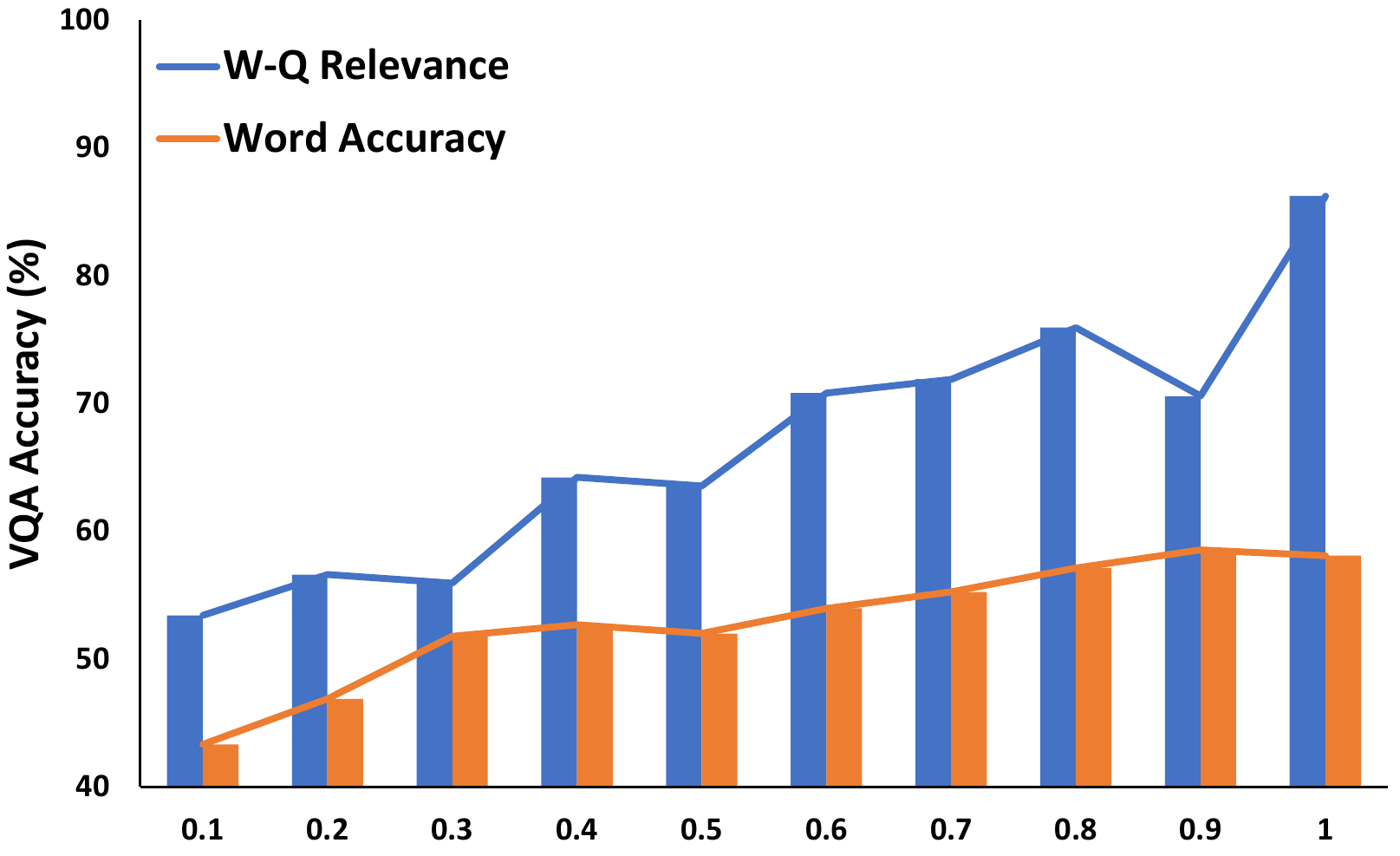}}
	\caption{The comparison of the impact of word accuracy and word-question relevance on VQA performance.}
	\label{fig:word-based VQA}
\end{figure}

An important characteristics of our framework is that the quality of explanations can influence the final VQA performance. In this section, we analyze the impact of the quality of predicted words on the VQA accuracy. We measure the quality from two sides: word accuracy and word-question relevance, illustrated in Eq.\ref{eq:word accuracy} and Eq.\ref{eq:w-q relevance}, respectively. And we normalize all the measurements into $[0,1]$.
Table \ref{tab:word_accuracy} shows the relationship between word accuracy and VQA performance. We can learn that the more accurate the predicted words, the better the VQA performance. This indicates that improving the visual detectors to predict more accurate words can benefit VQA performance.
Table \ref{tab:w-q relevance} shows the relationship between word-question relevance and VQA performance. Similar to word accuracy, the more relevant to the question the predicted words, the better the VQA performance. Particularly, when the word-question relevance exceeds 0.8, the predicted words are highly pertinent to the question, boosting the VQA accuracy to 76.15\%.
Based on the above observations, we conclude that high-quality word-level explanations can benefit the VQA performance a lot. 
Also, as shown in Fig.\ref{fig:word-based VQA}, word-question relevance has a bigger impact on the final VQA performance than word accuracy does.

\subsection{Sentence-based VQA}

\begin{table}[!htb] \vspace{-5mm}
	\caption{The relationship between sentence quality and VQA accuracy (\%).}
	\begin{subtable}{.5\linewidth}
		\centering
		\caption{Sentence accuracy}
		\label{tab:sentence_accuracy}%
		\begin{tabular}{cc}
			\toprule
			\begin{tabular}{@{}c@{}}Sentence \\ accuracy\end{tabular}
			& 
			\begin{tabular}{@{}c@{}}VQA \\ accuracy \end{tabular} \\
			\midrule
			\relax [0.0, 0.2) & 51.29 \\
			\relax [0.2, 0.8) & 55.53 \\
			\relax [0.8, 1.0) & 61.33 \\
			\bottomrule
		\end{tabular}
	\end{subtable}%
	\begin{subtable}{.5\linewidth}
		\centering
		\caption{S-Q Relevance}
		\label{tab:s-q relevance}%
		\begin{tabular}{cc}
			\toprule
			\begin{tabular}{@{}c@{}}S-Q \\ Relevance\end{tabular}
			& 
			\begin{tabular}{@{}c@{}}VQA \\ accuracy \end{tabular} \\
			\midrule
			\relax [0.0, 0.2) & 52.79 \\
			\relax [0.2, 0.8) & 62.34 \\
			\relax [0.8, 1.0) & 89.81 \\
			\bottomrule
		\end{tabular}
	\end{subtable} 
\end{table}

\begin{figure}[!tb] \vspace{-3mm}
	\centering {\includegraphics[width=0.45\textwidth]{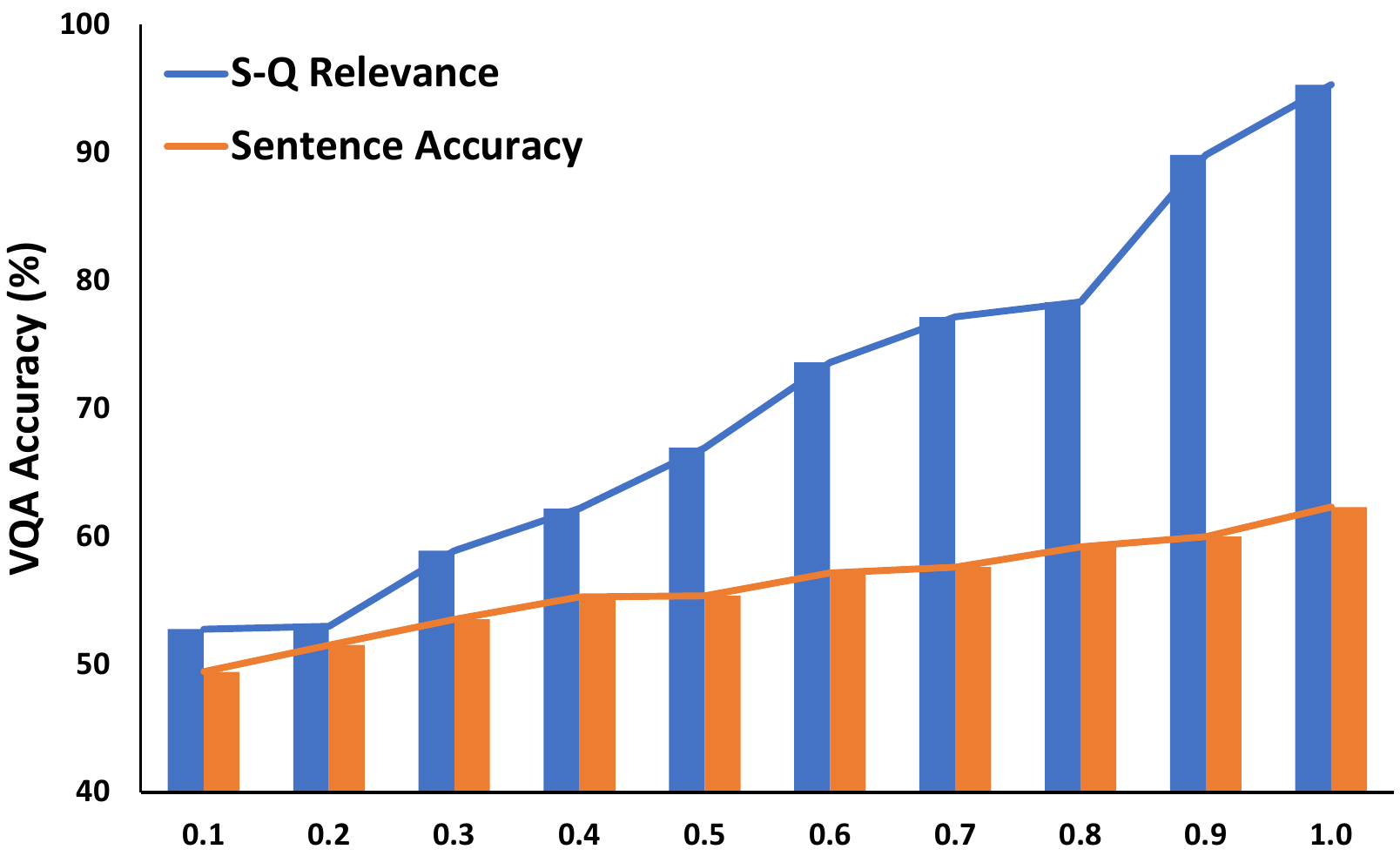}}
	\caption{The comparison of the impact of sentence accuracy and sentence-question relevance on VQA performance.}
	\label{fig:sentence-based VQA}
\end{figure}

In this section, we evaluate the sentence-based VQA model and analyze the relationship between the quality of the generated sentence and the final VQA performance.
Similar to the quality measurements of predicted words, we focus on the accuracy of the generated sentence itself and the relevance between sentence and question.
As shown in Table \ref{tab:sentence_accuracy}, the more accurate the generated sentence, the higher the VQA accuracy. The results suggest that the VQA performance can be further improved by a better image captioning model.
Table \ref{tab:s-q relevance} shows the impact of sentence-question relevance on VQA accuracy. We can see that the more relevant to the question the generated sentence, the better the VQA performance. Once the relevance reaches 0.8, the accuracy can significantly increase to 89.81\%. This proves that a sentence highly related to the question is more likely to contain the key information for the VQA module to answer the question.
Figure \ref{fig:sentence-based VQA} illustrates the comparison of the impact of sentence accuracy and sentence-question relevance on the final VQA performance. 
Sentence-question relevance has a greater influence on VQA performance than sentence accuracy does.

\begin{figure}[!tb] \vspace{-3mm}
	\centering {\includegraphics[width=0.5\textwidth]{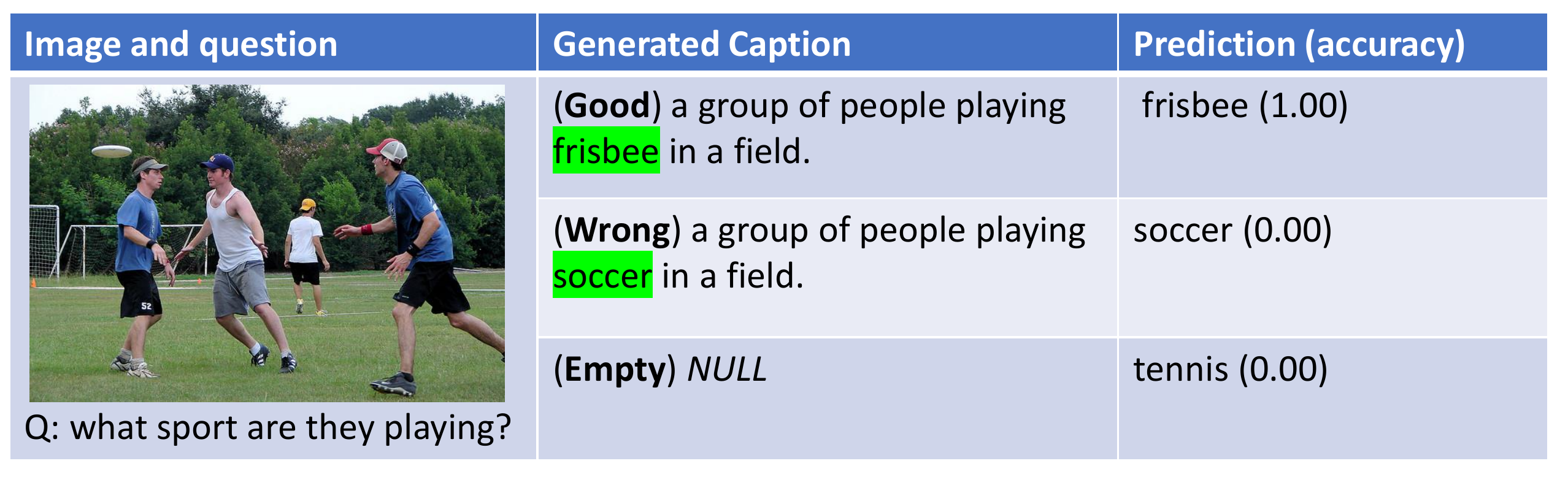}}
	\caption{A control case for comparing the accuracy when inputting captions of different quality. When getting a caption of high quality (the first one), the system can answer the question correctly. If we manually change the ``frisbee'' to ``soccer'', a wrong answer is predicted. When using an empty sentence, the system predicts the most popular answer ``tennis'' for this question.}
	\label{fig:sentence_control_case}
\end{figure}

To further verify the causal relationship between sentence quality and VQA performance, we conduct the following control experiments. First, we evaluate sentence-based VQA model when feeding different sources of captions with ascending quality: \textit{null} (only including an ``\#end'' token), \textit{sentence generation} and \textit{relevant groundtruth} (selecting from the groundtruth captions the most relevant one to the question). As shown in Table~\ref{tab:sent_verify}, \textit{sentence generation} performs much better than \textit{null}. And using \textit{relevant groundtruth} captions, the accuracy can improve by another 1.2 percent. Figure~\ref{fig:sentence_control_case} presents an example to illustrate the effect of the sentence quality on the accuracy.
From the above analysis, we can safely reach the conclusion that the VQA performance can be greatly improved by generating sentence-level explanations of high quality, especially of high relevance to the question.

\begin{table}[htbp]
	\centering
	\caption{Performance comparison on the validation split of VQA-real open-ended task when the sentence-based VQA model uses different sources of captions. (accuray in \%)}
    \begin{tabular}{lcccc}
	\toprule
	\multicolumn{1}{c}{\multirow{2}[4]{*}{Caption source}} & \multicolumn{4}{c}{validation} \\
	\cmidrule{2-5}          & All   & Y/N   & Num   & Others \\
	\midrule
	null  & 46.21 & 73.82 & 34.98 & 25.63 \\
	sentence generation  & 54.85 & 76.31 & 36.64 & 42.23 \\
	relevant groundtruth  & 56.05 & 77.42 & 41.04 & 44.34 \\
	\bottomrule
	\end{tabular}%
	\label{tab:sent_verify}%
\end{table}%

\subsection{Case Study}
\begin{figure*}[!tb] \vspace{-10mm}
	\centering {\includegraphics[width=1\textwidth]{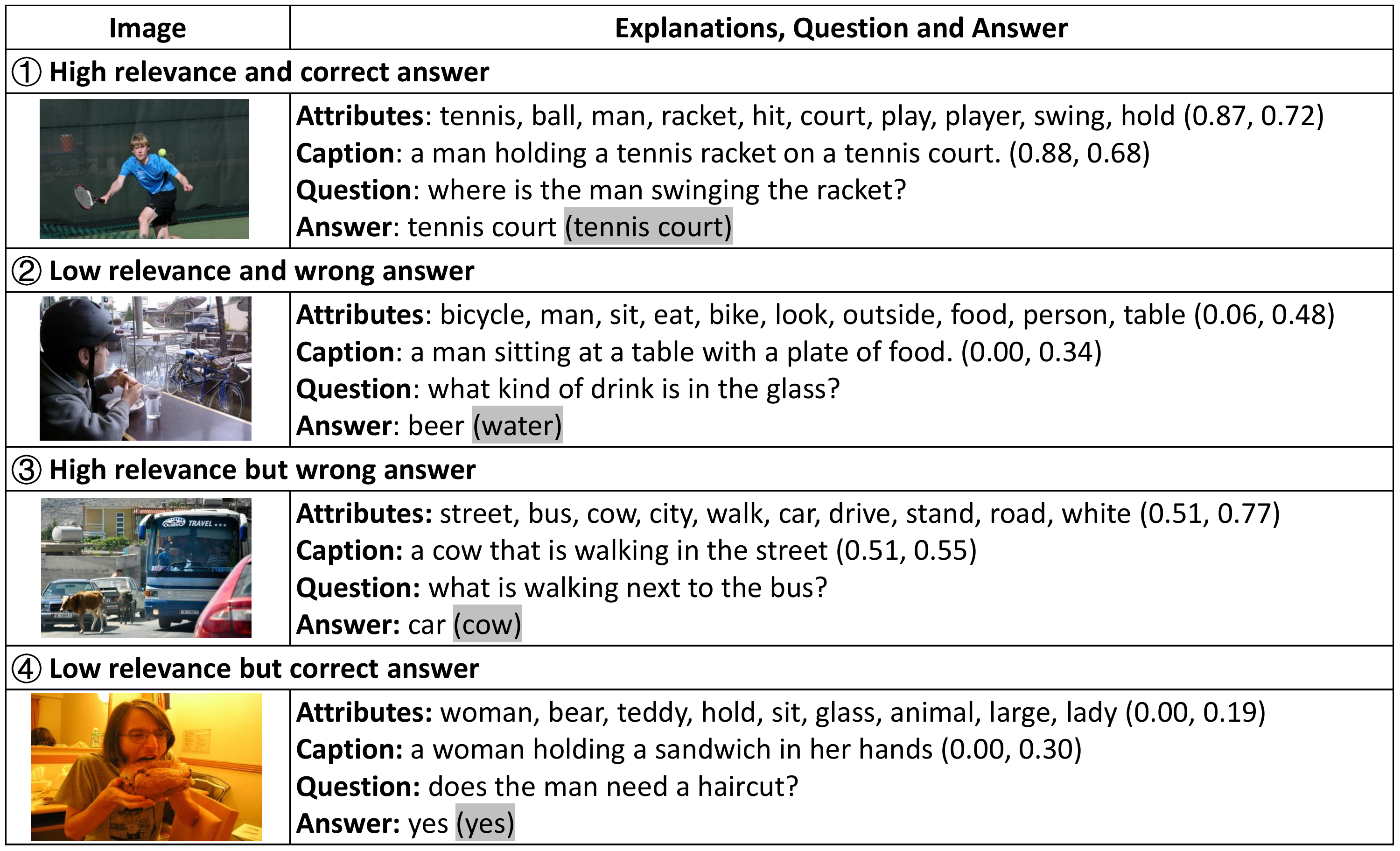}}
	\caption{Four types of cases in our results: 1). high relevance and correct answer; 2). low relevance and wrong answer; 3). high relevance but wrong answer; 4). low relevance but correct answer. ``(*,*)'' behind the explanations (attributes/caption) denotes the explanation-question relevance score and explanation accuracy, respectively. \colorbox[rgb]{ .682,  .667,  .667}{Gray} denotes groundtruth answers.}
	\label{fig:case_study}
\end{figure*}

\begin{figure*}[!tb] \vspace{-3mm}
	\centering {\includegraphics[width=1\textwidth]{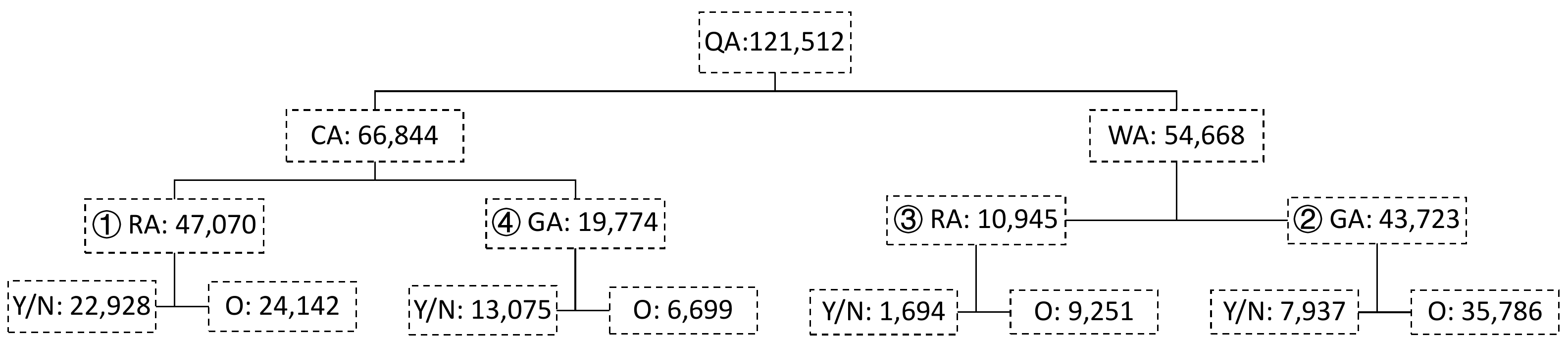}}
	\caption{Dataset dissection according to the four types of cases. We define that the answer is guessed when the explanations are irrelevant to the question and otherwise reliable. The case numbers in the third row correspond to these in Fig.\ref{fig:case_study}.
		QA: all questions and answers.
		CA: questions with correct answers.
		WA: questions with wrong answers.
		GA: questions with guessed answers.
		RA: questions with reliable answers.
		Y/N: answer type ``yes/no''.
		O: answer types other than ``yes/no''.
	}
	\label{fig:dataset_analysis}
\end{figure*}

From the above evaluation of word-based and sentence-based VQA model, we conclude that the relevance between explanations (attributes/caption) and the question has a great impact on the final VQA performance. In this section we illustrate this conclusion by studying four possible types of cases: 1). \textit{high relevance and correct answer}; 2). \textit{low relevance and wrong answer}; 3). \textit{high relevance but wrong answer}; 4). \textit{low relevance but correct answer}.

\noindent\textbf{High relevance and correct answer}. From the first case in Fig. \ref{fig:case_study}, we can see that the explanations for the image are highly relevant to the question: both the predicted attributes and the generated sentence contain the words ``man'' and ``racket'' occurring in the question. And the explanations also has key information that can predict the answer ``tennis court.'' In this type of case, the system successfully extracts from the image the relevant information that covers the question, facilitating answer generation.

\noindent\textbf{Low relevance and wrong answer}. In the second case, although the attributes and caption can reflect part of the image content such as ``man'' and ``food'', they neglect the key information about the ``glass'' that is asked in the question. The absence of ``glass'' in the explanations produces a low explanation-question relevance score and leads the system to a wrong answer. 
In this type of case, two lessons can be derived from the low relevance:
1). as the explanations are irrelevant to the question, the system tends to predict the most frequent answer (``beer'') for this question type (``what kind of drink ...''), which implies that the answer is actually \textit{guessed} from the dataset bias;
2). the error comes from the image understanding part rather than the question answering module, because the system fails to extract from the image enough information to answer the question in the first place. 
This error suggests that some improvements are needed in word prediction and sentence generation modules to generate more comprehensive explanations for the image.

\noindent\textbf{High relevance but wrong answer}. In the third case, we can see that although the system fails to predict the correct answer, the explanations for the image are indeed relevant to the question and the system also recognize the key information ``cow.'' This indicates that the error is caused by the question answering module rather than the explanation generation part. The system can recognize that ``a cow is walking in the street'' and ``a bus is in the street'', but it fails to conclude that ``the cow is next to the bus.'' This error may lie in the weakness of LSTM which struggles on such complex spatial relationship inference. In the following analysis, we would show that such cases only occupy a relatively small proportion of the whole dataset.

\noindent\textbf{Low relevance but correct answer}. In the last example of Fig.~\ref{fig:case_study}, we know from the explanations that the system mistakes the ``man'' in the image for ``woman'' and neglects the information about his ``hair.'' The explanations, therefore, have a low relevance score, which indicates that the answer ``yes'' is guessed by the system. Although the guessed answer is correct, it cannot be credited to the correctness of the system. In fact, for this particular answer type ``yes/no'', the system has at least 50\% chance to hit the right answer.

We dissect all the results in the dataset according to the above four types of cases, as shown in Fig.~\ref{fig:dataset_analysis}. Among the questions that the system answers correctly, nearly 30\% are guessed. This discovery indicates that, buried in the seemingly promising performance, the system actually takes advantage of the dataset bias, rather than truly understands the image content. 
Over 65\% of the answers that are correctly guessed belong to ``yes/no'', an answer type easier for the system to hit the right answer than other types.
As for the questions to which the system predicts wrong answers, a large proportion (around 80\%) has a low explanation-question relevance, which means that more efforts need to be put into improving the attributes detectors and image captioning model. 
Questions with other answer types account for more than 80\% of the wrongly-guessed answers. This is not surprising because for these questions the system cannot rely on the dataset bias anymore, considering the great variety of the candidate answers.

\subsection{Performance Comparison}
\begin{table}[htbp] \vspace{-3mm}
	\centering
	\caption{Performance comparison with the state-of-the-art. We show the performance on both test-dev and test-standard splits of VQA-real open-ended task. The performances are achieved by training the VQA module on both the train and val splits. \colorbox[rgb]{ .682,  .667,  .667}{MCB-ensemble} incorporates external training data, which is an \textit{extra} advantage over other methods. \colorbox[rgb]{ .682,  .667,  .667}{Human} is the human performance for reference.}
	\resizebox{1\columnwidth}{!}{
		\begin{tabular}{ccccccccc}
			\toprule
			\multirow{2}[4]{*}{Method} & \multicolumn{4}{c}{test-dev}  & \multicolumn{4}{c}{test-standard} \\
			\cmidrule{2-9}          & All   & Y/N   & Num   & Others & All   & Y/N   & Num   & Others \\
			\midrule
			LSTM Q+I \cite{antol2015vqa} & 53.74 & 78.94 & 35.24 & 36.42 & 54.06 & 79.01 & 35.55 & 36.80 \\
			Concepts \cite{wu2016value} & 57.46 & 79.77 & 36.79 & 43.10 & 57.62 & 79.72 & 36.04 & 43.44 \\
			ACK \cite{wu2016ask} & 59.17 & 81.01 & 38.42 & 45.23 & 59.44 & 81.07 & 37.12 & 45.83 \\
			SAN \cite{yang2016stacked} & 58.70 & 79.30 & 36.60 & 46.10 & 58.90 & -     & -     & - \\
			HieCoAtt \cite{lu2016hierarchical} & 61.80 & 79.70 & 38.70 & 51.70 & 62.10 & -     & -     & - \\
			MCB \cite{fukui2016multimodal} & 60.80 & 81.20 & 35.10 & 49.30 & -     & -     & -     & - \\
			\rowcolor[rgb]{ .682,  .667,  .667} MCB-ensemble \cite{fukui2016multimodal} & 66.70 & 83.40 & 39.80 & 58.50 & 66.50 & 83.20 & 39.50 & 58.00 \\
			\rowcolor[rgb]{ .682,  .667,  .667} Human \cite{antol2015vqa} & -     & -     & -     & -     & 83.30 & 95.77 & 83.39 & 72.67 \\
			\midrule
			Word-based VQA & 56.76 & 77.57 & 35.21 & 43.85 & -     & -     & -     & - \\
			Sentence-based VQA & 57.91 & 78.03 & 36.73 & 45.52 & -     & -     & -     & - \\
			Full VQA & 59.93 & 79.32 & 38.41 & 48.25 & 60.07 & 79.09 & 38.25 & 48.57 \\
			\bottomrule
	\end{tabular}}
	\label{tab:performance}%
\end{table}%

In this section, we present the performance comparison between variants of our framework and the state-of-the-art.  
For a fair comparison, all the results are reported on the test-dev and test-standard splits of which ground-truth answers are not released and all the performances are returned by the test server \cite{antol2015vqa}. 
From Table \ref{tab:performance}, we can see that sentence-based VQA consistently outperforms word-based VQA, which indicates that sentence-level explanations are superior to word-level ones. This is because the generated captions not only include the objects in the image, but also encode the relationship between these objects, which is important for predicting the correct answer.
Moreover, full VQA model obtains a better performance by combining attributes and captions.
As the answer reasoning module in our framework is based on explanations, if we can get better attribute detectors and image captioning model, the VQA performance can be further improved.

Compared with the state-of-the-art, our framework achieves better performance than LSTM Q+I \cite{antol2015vqa}, Concepts \cite{wu2016value}, and ACK \cite{wu2016ask}, which use CNN features, high-level concepts, and external knowledge, respectively. 
Both SAN \cite{yang2016stacked} and HieCoAtt \cite{lu2016hierarchical} use attention mechanism in images or questions and yield comparable performance with ours.
MCB-ensemble \cite{fukui2016multimodal} achieves significantly better performance than ours and other top methods, but it suffers from a high-dimensional feature (16,000 vs 1,280), which poses a limitation on the model's efficiency.
The main advantage of our framework over other methods is that it not only predicts an answer to the question, but also generates human-readable attributes and captions to explain the answer. These explanations can help us understand what the system extracts from an image and their relevance to the question. As explanations improve, so would our system.

\section{Discussions and Conclusions} \label{sec:CO}
In this work, we break up the end-to-end VQA pipeline into \textbf{explaining} and \textbf{reasoning}, and achieve comparable performance with the state-of-the-art.
Different from previous work, our method first generates attributes and captions as explanations for an image and then feed these explanations to a question answering module to infer an answer. The merit of our method lies in that these attributes and captions allow a peek into the process of visual question answering. Furthermore, the relevance between these explanations and the question can act as indication whether the system really understands the image content. 

It is worth noting although we also use the CNN-RNN combination, we generate words and captions as the explanations of images, thus allowing the VQA system to perform {\it reasoning on semantics} instead of unexplanable CNN features. Since the effectiveness of CNN for generating attributes and captions is well established, the use of CNN as a component does not contradict our high-level objective for explanable VQA. Our goal is not to immediately make a big gain in performance, but to propose a more powerful framework for VQA. Our current implementation already matches the state of the art, but more importantly, provides to the ability to explain and to improve.


{\small
\bibliographystyle{ieee}
\bibliography{egbib}
}

\end{document}